\documentclass[11pt,a4paper]{article}
\usepackage[hyperref]{emnlp-ijcnlp-2019}
\usepackage{times}
\usepackage{latexsym}

\usepackage{url}

\usepackage{enumerate}
\usepackage{url}
\usepackage{mathrsfs}
\usepackage{amsmath}
\usepackage{amssymb}
\usepackage{verbatim}
\usepackage{graphicx}     
\usepackage{subfigure}
\usepackage{bm}
\newenvironment{sequation}{\begin{equation}\small}{\end{equation}}

\aclfinalcopy 

\title{Self-Balanced Dropout}

\author{
  Shen Li\textsuperscript{$^{\spadesuit}$} \And
  Chenhao Su\textsuperscript{$^{\clubsuit}$} \And
  Renfen Hu\textsuperscript{$^{\heartsuit}$} \And
  Zhengdong Lu\textsuperscript{$^{\spadesuit}$} \AND
  $^{\spadesuit}${\tt \{shen, luz\}@deeplycurious.ai} \\
  $^{\clubsuit}${\tt suchenhao821@bupt.edu.cn} \\
  $^{\heartsuit}${\tt irishu@mail.bnu.edu.cn} \\
  $^{\spadesuit}$ Deeplycurious.ai \\
  $^{\clubsuit}$ SICE, Beijing University of Posts and Telecommunications \\
  $^{\heartsuit}$ Institute of Chinese Information Processing, Beijing Normal University \\
}

\date{}

\begin{document}
\maketitle
\begin{abstract}
Dropout is known as an effective way to reduce overfitting via preventing co-adaptations of units. 
In this paper, we theoretically prove that the co-adaptation problem still exists after using dropout due to the correlations among the inputs.
Based on the proof, we further propose Self-Balanced Dropout, a novel dropout method which uses a trainable variable to balance the influence of the input correlation on parameter update. 
We evaluate Self-Balanced Dropout on a range of tasks with both simple and complex models. 
The experimental results show that the mechanism can effectively solve the co-adaption problem to some extent and significantly improve the performance on all tasks. \footnote{Source codes are released at \href{https://github.com/shenshen-hungry/Self-Balanced-Dropout}{https://github.com/shenshen-hungry/Self-Balanced-Dropout}.}
\end{abstract}

\section{Introduction}

Dropout~\cite{hinton2012improving,srivastava2014dropout}, an effective algorithm to reduce overfitting, has been widely used in the training of neural networks. 
The key idea is to randomly drop out units (input or hidden layer units) of a neural network during training. 
Dropout can be seen as a kind of regularization~\cite{wager2013dropout,baldi2013understanding,srivastava2014dropout,helmbold2015inductive}. 

In this paper, we find that the co-adaptation problem still exists when the input has a strong correlation, and it will cause a certain degree of overfitting.
An intuitive example is in many natural language processing (NLP) tasks, words are the inputs of a model. 
The word distribution hypothesis~\cite{firth1957synopsis} states that a word is characterized by the company it keeps. 
Therefore, for a sentence, there are some natural correlations between the words, such as \emph{cat} and \emph{mouse}, \emph{apple} and \emph{eat}. 
This may lead to the co-adaptations of units which can work well on a training set but cannot generalize to unseen data.
Although dropout seems to prevent the co-adaptations by randomly dropping units during training, the problem still remains due to the accumulation of parameter updates.

Based on the theoretical analysis, we propose Self-Balanced Dropout, a simple but effective method that solves the co-adaptation problem caused by the input correlation.
Different from the original dropout which randomly sets units to 0, this method randomly replaces units with trainable variables at each iteration.
These trainable variables can reduce the impact of correlation among inputs on parameter update, allowing parameters to be updated properly.
The experimental results show that Self-Balanced Dropout consistently improves the performance over the original dropout in various NLP tasks.

\section{Motivation}

~\newcite{srivastava2014dropout} propose the dropout mechanism, and illustrate how dropout works as a regularization term in linear models. 
In this section, we will brifely introduce their proof and further address the problem on parameter updates caused by correlation of the inputs.

Specifically, let $X=(x_1, x_2, ..., x_n)^T \in \mathbb{R}^{N \times D}$ be a data matrix, where each $x_i \in \mathbb{R}^{D}$ represents a D-dimensional data sample. 
$y\in \mathbb{R}^{N}$ be the label of the data. Linear regression tries to find a $w\in \mathbb{R}^{D}$ that minimizes

\begin{sequation}
    J(w)=\left \|y-Xw \right \|^2.
\end{sequation}
\vspace{-0.3cm}

Dropout algorithm randomly perturbs the features of the inputs. 
For every modified input sample $\tilde{x}_i$, $x_{ij}$ is maintained with the keep probability $p$, or set to 0 with probability $1-p$.
In practice, weight scaling is used to keep training and testing consistent, where each $x_{ij}$ is scaled down by $p$ during training. 
Then the final $\tilde{x}_{ij}$ is
\begin{sequation}
    \tilde{x}_{ij}=\begin{cases}x_{ij}/p, &with \ probability \ p \cr 0, &with \  probability \ q=1-p\end{cases}
\end{sequation}

After applying dropout, the input data matrix can be expressed as $R \circ \tilde{X}$, where $R \in \{ 0,1 \}^{N \times D}$ is a random matrix with $r_{ij} \sim Bernoulli ( p )$ , $\tilde{X}$ is the scaled data matrix and $\circ$ denotes an element-wise product. Then the objective function becomes
\begin{sequation}
\begin{split}
\underset{w}{minimize}\;\mathbb{E}_{R\sim Bernoulli\left ( p \right )}\left [  \left \|y-(R \circ \tilde{X})w \right \|^2\right ].
\end{split}
\end{sequation}
This reduces to
\begin{sequation}
\label{formula: target in original dropout}
\begin{split}
\underset{w}{minimize}\; \left \|y-Xw \right \|^2 +R(w).
\end{split}
\end{sequation}
where $R(w)= \frac{1-p}{p}\sum_{i}\sum_{j}(x_{ij}w_j)^2$. 
As we can see, the role of dropout in the linear regression is equivalent to a regular term that depends on the dropout probability $p$ and the inputs $x$.

Further, the parameter $w$ is updated by the standard stochastic gradient descent method with a learning rate of $\alpha$ like
\begin{sequation}
\begin{split}
w_j:=w_j-\alpha (\frac{\partial J(w)}{\partial w_j} + \frac{\partial R(w)}{\partial w_j}),
\end{split}
\end{sequation}
where
\begin{sequation}
\begin{split}
\frac{\partial R(w)}{\partial w_j}= 2\frac{1-p}{p}  \sum_{i}x_{ij}^{2}w_j.
\end{split}
\end{sequation}
When $w$ is small and $x$ is large, the parameter update in the regularization term will highly depend on $x$. 
However, the update direction of $w$ should be guided by the label $y$, rather than the input $x$. 
When some features in $x$ are highly correlated, the regularization term will make the corresponding dimensions of $w$ be updated in a similar direction. 
This will still lead to the co-adaptation problem of weights with which the model may not work well when the inputs are not highly correlated at testing.

\section{Self-Balanced Dropout}

To solve the problem mentioned above, we propose Self-Balanced Dropout, to randomly replace the units of the inputs with a trainable variable \emph{$x_{mask}$}. 
Then the parameter update will not be excessively affected by the highly correlated input features, thus alleviating co-adaptation.

Formally, when Self-Balanced Dropout is applied, $\tilde{x}_{ij}$ is expressed as
\begin{sequation}
    \tilde{x}_{ij}=\begin{cases}x_{ij}, &with \ probability \ p \cr x_{mask}, &with \  probability \ q=1-p\end{cases}
\end{sequation}

Since we have not zeroed any unit, all the units can emit information to the next layer and thus scaling is not needed. 
Then the objective function becomes
\begin{sequation}
    \begin{split}
    \underset{w}{minimize}\quad\mathbb{E}_{R\sim Bernoulli\left ( p \right )} [   \|y-(R \circ X)w\\- [(I-R) \circ X_{mask}]w\|^2].
    \end{split}
\end{sequation}
This reduces to 
\begin{sequation}
\label{formula: target in self-balanced dropout}
\begin{split}
\underset{w}{minimize}\quad \left \|y-pXw \right \|^2 +Q(w)+\hat{R}(w).
\end{split}
\end{sequation}
where
\begin{sequation}
\begin{aligned}
Q(w)&= \left\|y-(1-p)X_{mask}w \right \|^2,\\
\hat{R}(w)&= p(1-p)\sum_{i}\sum_{j}[(x_{ij}+x_{mask})w]^2,
\end{aligned}
\end{sequation}
where $X_{mask} \in \mathbb{R}^{N\times D}$ is the mask matrix in which each value is the trainable variable $x_{mask}$. 
Comparing the formula~\ref{formula: target in original dropout} with the formula~\ref{formula: target in self-balanced dropout}, Self-Balanced Dropout brings two changes. 
Firstly, $Q(w)$ forces $w$ and $x_{mask}$ to be in an inverse relationship, i.e. when $w$ is small, $x_{mask}$ will be large. 
Secondly, it should be noted that
\begin{sequation}
\frac{\partial \hat{R}(w)}{\partial w_j} = 2p(1-p)\sum_{i}(x _{ij}+x_{mask})^{2}w_j.
\end{sequation}
The update direction of $w$ is now determined by both $x_{ij}$ and $x_{mask}$. 
It is not hard to derive that adding a large $x_{mask}$ to $x$ can balance the effect of $x$ on $w$, which diminish the influence of co-adaptation.

It should be noted that some methods~\cite{vincent2008extracting,devlin2018bert} randomly mask or replace a few of the tokens at the input layer to force the model to keep a contextual representation of every input token. 
It is only an empirical method and no one knows the reason why it works well.
From the viewpoint of the above proof, the replacing method is actually an improved dropout in the input layer and it can further reduce the co-adaptation that cannot be prevented by the original dropout. 
Obviously, the co-adaptation problem exists not only in the input layer but also in the hidden layers.
Self-Balanced Dropout is not limited to the input layers and can be applied to any layer of a model including both input layers and hidden layers.

\section{Experiment}
\begin{figure}[ht]
\centering
\includegraphics[scale=0.2]{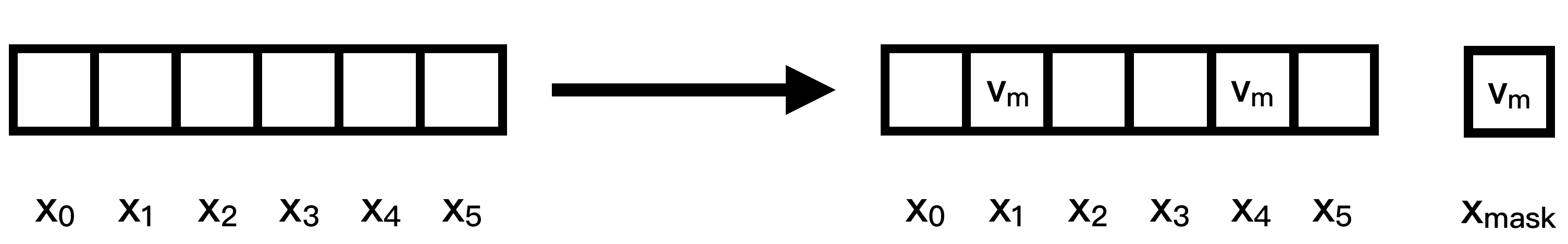}
\caption{mask with a trainable value}
\label{fig: value}
\end{figure}

\begin{figure}[ht]
\centering
\includegraphics[scale=0.2]{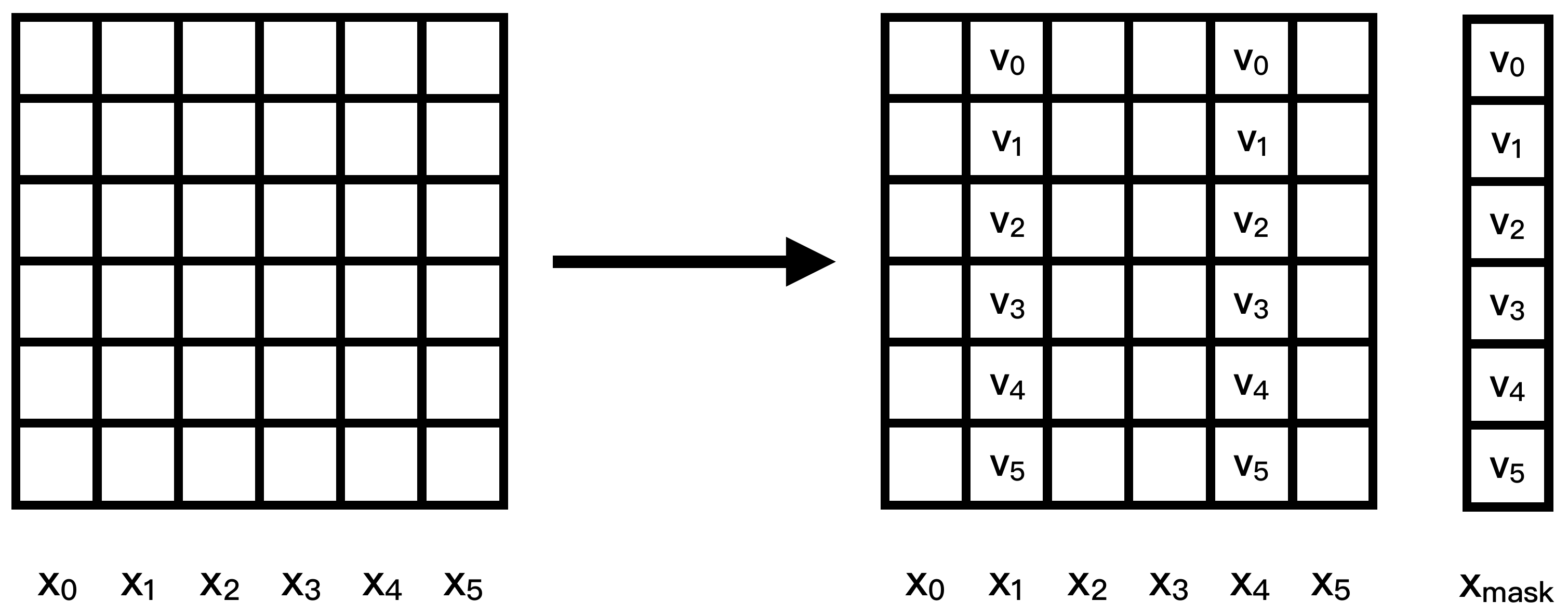}
\caption{mask with a trainable vector}
\label{fig: vector}
\end{figure}

We evaluate Self-Balanced Dropout on three tasks: text classification, named entity recognition (NER) and machine translation. 
For each input $x_i$, the mask variable is a trainable value (Figure~\ref{fig: value}), if $x_i$ is a value, and the mask variable is a trainable vector (Figure~\ref{fig: vector}), if $x_i$ is a vector, e.g. $x_i$ is a word representation.

\vspace{-0.1cm}

\vspace{-0.1cm}

\begin{table*}
\small
\centering
    \begin{tabular}{cccccccc}
    \hline
    Model          &\hspace{-0.3cm} MR            & SST-1         & SST-2          & Subj          & TREC          & CR            & MPQA \\
    \hline
    CNN-non-static &\hspace{-0.3cm} 81.5          & 48.0          & 87.2           & 93.4          & 93.6          & 84.3          & 89.5 \\
    CNN-SB-Dropout &\hspace{-0.3cm} \textbf{81.7} & \textbf{52.0} & \textbf{88.8}  & \textbf{94.0} & \textbf{94.2} & \textbf{85.0} & \textbf{90.0}\\ 
    ($p_1$, $p_2$) &\hspace{-0.3cm} (0.8, 0.4)    & (0.9, 0.7)    & (1.0, 0.6)     & (0.5, 0.6)    & (0.8, 0.6)    & (0.9, 0.6)    & (0.9, 0.4) \\
    \hline
    MGNC-CNN       &\hspace{-0.3cm} -             & 48.7          & 88.3           & \textbf{94.1} & 95.5          & -             & -  \\ 
    MVCNN          &\hspace{-0.3cm} -             & 49.6          & \textbf{89.4}  & 93.9          & -             & -             & -  \\ 
    DSCNN          &\hspace{-0.3cm} \textbf{82.2} & 50.6          & 88.7           & 93.9          & \textbf{95.6} & -             & -  \\ 
    Semantic-CNN   &\hspace{-0.3cm} 82.1          & \textbf{50.8} & 89.0           & 93.7          & 94.4          & \textbf{86.0} & 89.3  \\ 
    TopCNN${\rm _{word}}$ &\hspace{-0.3cm} 81.7   & -             & -              & 93.4          & 92.5          & 84.9          & \textbf{89.9}  \\
    \hline
    \end{tabular}
    \caption{\label{1}Effectiveness of Self-Balanced Dropout on sentence classification task. 
    $\mathbf{p_1}$ and $\mathbf{p_2}$ are the keep probability (1 - dropout rate) of the input layer and the hidden layer respectively.
    The first line is the result of CNN-non-static model in~\cite{kim2014convolutional}. 
    Results also include: MGNC-CNN~\cite{zhang2016mgnc}, MVCNN~\cite{yin2016multichannel},
    DSCNN~\cite{zhang2016dependency}, Semantic-CNN~\cite{li2017initializing} and TopCNN${\rm _{word}}$~\cite{zhao2017topic}.}
\end{table*}

\subsection{Datasets and Experiment Settings}

\subsubsection{Sentence Classification} 

We employ the same seven datesets with~\cite{kim2014convolutional}, including both sentiment analysis and topic classification tasks.
\textbf{MR}: Movie reviews sentiment datasets~\cite{pang2005seeing}.
\textbf{SST-1}: Stanford Sentiment Treebank with 5 sentiment labels~\cite{socher2013recursive}.
To keep same with~\cite{kim2014convolutional}, we train the model on both phrases and sentences but only test on sentences.
\textbf{SST-2}: SST-1 data with binary labels.
\textbf{Subj}: Subjective or objective classification dataset~\cite{pang2004sentimental}.
\textbf{TREC}: 6-class question classification dataset~\cite{li2002learning}.
\textbf{CR}: Customer products review dataset~\cite{hu2004mining}.
\textbf{MPQA}: Opinion polarity dataset~\cite{wiebe2005annotating}.

CNN-non-static proposed by~\newcite{kim2014convolutional} is used as our baseline. 
We replace the original dropout before the fully connected layer with Self-Balanced Dropout like Figure~\ref{fig: value}. 
In addition, ~\newcite{zhang2015sensitivity} mention that dropout at the input layer helps little. 
Therefore, in the baseline, the input layer does not use dropout, which is equivalent to setting the keep probability to 1. 
Considering high correlation among the inputs may exsit because of the word distribution hypothesis, thus we assume that it is suitable to use Self-Balanced Dropout at the first layer. 
In the input layer, every word embedding can be replaced with the mask variable according to its Self-Balanced Dropout probability like Figure~\ref{fig: vector}. 
For a fair comparison, we use the same hyper-parameter setting with ~\newcite{kim2014convolutional}'s work.

\subsubsection{Named Entity Recognition}

We test Self-Balanced Dropout on the CoNLL-2003~\cite{tjong2003introduction} and the OntoNotes 5.0~\cite{hovy2006ontonotes, pradhan2013towards}. 
A strong baseline ID-CNN~\cite{strubell2017fast} is chosen to be our baseline.
Similarly, we replace the original dropout between each convolutional layer of ID-CNN with Self-Balanced Dropout like Figure~\ref{fig: vector}. 
Except for the dropout module, the rest of the model is consistent with the baseline.

\subsubsection{Machine Translation}

We also replace the original dropout with Self-Balanced Dropout in Transformer, an influential deep model in NLP.
Transformer-base model with the same setting in ~\cite{vaswani2017attention} is used, except for the Self-Balanced Dropout rate which is 0.05 smaller than the original one. \footnote{The experiments of translation are conducted in Tensor2Tensor~\cite{tensor2tensor}.}
We test the model on WMT 2014 English-German dataset.

\begin{figure}[ht]
\centering
\subfigure[]{
\label{Fig.sub.1}
\includegraphics[width=0.35\textwidth]{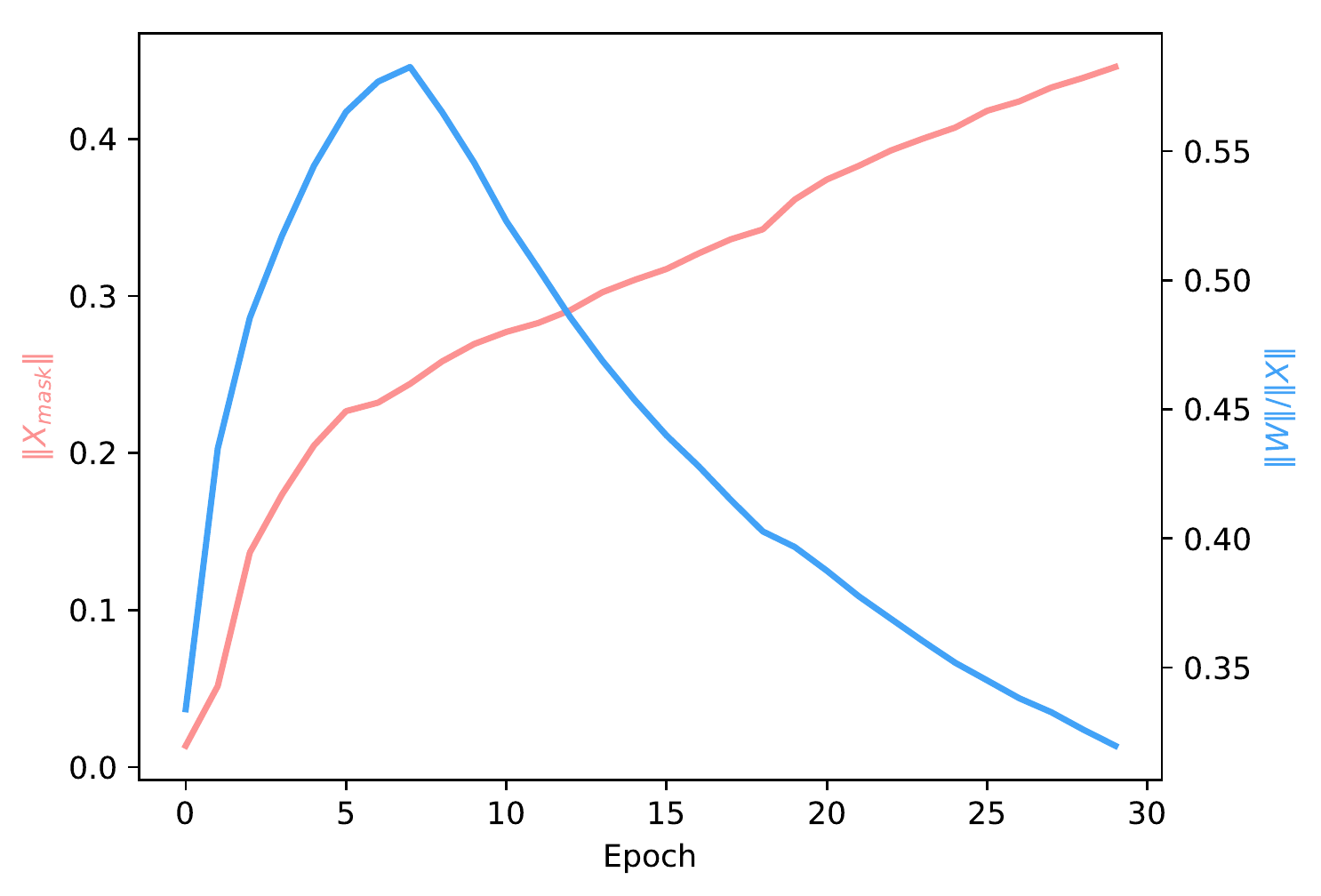}
}
\subfigure[]{
\label{Fig.sub.2}
\includegraphics[width=0.35\textwidth]{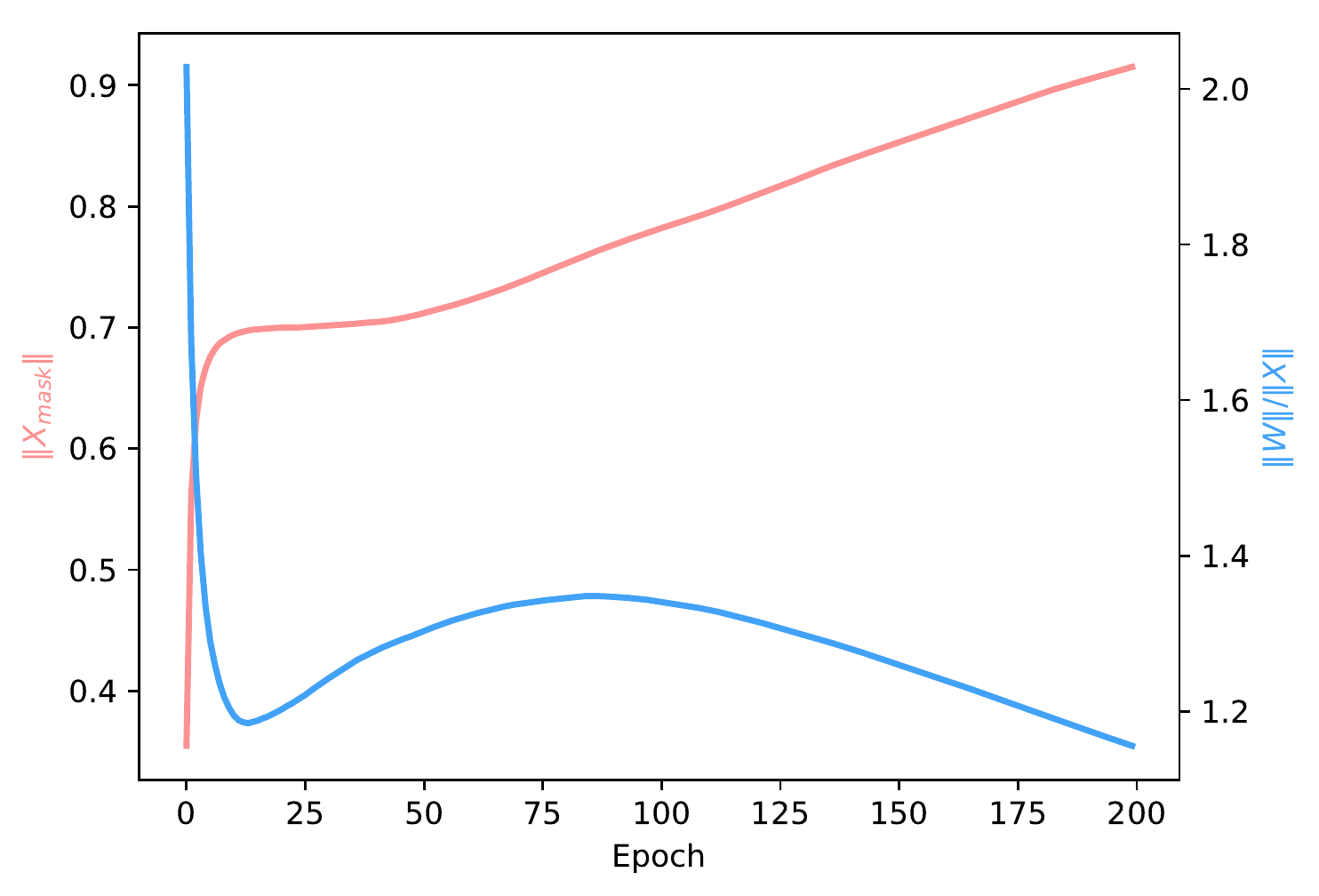}
}
\caption{Changes in the $\left \| W \right \|$/$\left \| X \right \|$ and $||X_{mask}||$ during training in two experiments. Figure~\ref{Fig.sub.1}, Figure~\ref{Fig.sub.2} are "CNN-hidden-layer" in MR and "IDCNN-hidden-layer" in CoNLL-2003 respectively.}
\label{hidden}
\end{figure}

\begin{table}[ht]
\small
\centering
    \begin{tabular}{cc}
    \hline
    Model          & F1    \\ \hline
    ID-CNN         & 90.32 $\pm$ 0.26  \\ 
    ID-CNN-SB-Dropout & \textbf{90.73} $\pm$ \textbf{0.25}  \\\hline
    \end{tabular}
    \caption{\label{2} F1 scores of models on CoNLL-2003.}
\end{table}

\begin{table}[ht]
\small
\centering
    \begin{tabular}{cc}
    \hline
    Model                      & F1    \\ \hline
    ID-CNN (3 blocks)          &\hspace{-0.4cm} 85.27 $\pm$ 0.24  \\
    ID-CNN-SB-Dropout (3 blocks) &\hspace{-0.4cm} \textbf{85.68} $\pm$ \textbf{0.20} \\ \hline
    \end{tabular}
    \caption{\label{3} F1 scores of models on OntoNotes 5.0. }
\vspace{-0.2cm}
\end{table}

\begin{table}[ht]
\small
\centering
    \begin{tabular}{cc}
    \hline
    Model                      &\hspace{-0.4cm} BLEU (case-sensitive)   \\ \hline
    Transformer-base          &\hspace{-0.4cm}  27.3 \\
    Transformer-base-SB-Dropout &\hspace{-0.4cm} \textbf{27.5} \\ \hline
    \end{tabular}
    \caption{\label{4} BLEU scores on WMT 2014 En-De dataset. }
\vspace{-0.2cm}
\end{table}

\subsection{Experimental Results and Analysis}

For sentence classification task, Table~\ref{1} shows the classification accuracies on seven datasets. 
The results of Self-Balanced Dropout are listed in ``CNN-SB-Dropout'' row. 
The revised dropout further improves accuracy on all seven datasets. 
We also list the results of other models which either use multiple pre-trained embedding as inputs or use more complex deep models. 
By adding $x_{mask}$ into the original dropout, Self-Balanced Dropout can also achieve competitive performance against these more sophisticated methods.

For NER task, Table~\ref{2} lists F1 scores of models on CoNLL-2003 and Table~\ref{3} lists F1 scores of models on OntoNotes 5.0. 
We list the result of Self-Balanced Dropout in "ID-CNN-SB-Dropout" row. 
Experiments show that the revised dropout consistently improves the performance over the baseline.

For translation task, Table~\ref{4} lists BLEU scores of Transformer models. 
Although Transformer has a very deep structure and a large number of parameters, Self-Balanced Dropout works effectively and gains an improvement.

In order to understand where the improvement comes from and to verify that our calculation is correct, we further analyze the changes in the norm of $X$, $X_{mask}$ and $w$ in the training time. 
Figure~\ref{hidden} shows the changes of $norm(w)/norm(X)$ and $X_{mask}$ in two experiments. 
After several epochs, the $norm(w)/norm(X)$ continues to decrease, which means the regularization term $R(w)$ is more determined by $X$. Meanwhile, $X_{mask}$ keeps increasing as expected, thus diminishes the influence of correlation in $X$ on parameter updates.

It is worth noting that in the above experiments, Self-Balanced Dropout is applied in a single dimension. 
We have tried to apply the method simultaneously in different dimensions, i.e. units in any dimension are randomly replaced with the same trainable variable. 
It is similar to the original dropout, while the improvement does not seem statistically significant.
The reason could be that sharing the same variable in different dimensions is meaningless, and it makes the trainable variable overburdened. 
Therefore, replacing inputs in a single dimension is better than that in different dimensions.

\section{Conclusion}

In this work, we identify the inherent problem with the original dropout in causing co-adaptation problem from the perspective of regularization. 
This motivates us to propose Self-Balanced Dropout, a method that aims to diminish the influence of correlation in inputs on parameter updates. 
Since our approach is to improve the original dropout, it can also be replaced with Self-Balanced Dropout wherever the original dropout is used.
The experimental results provide impressive improvements on all tasks.
In the future, we will extend the proposed method to other fields.

\bibliography{emnlp-ijcnlp-2019}
\bibliographystyle{acl_natbib}

\end{document}